# Integrated Image-Text Based on Semi-supervised Learning for Small Sample Instance Segmentation

Ruting Chi[a], Zhiyi Huang[a] and Yuexing Han[a,b,*]

[a]School of Computer Engineering and Science, Shanghai University, 99 Shangda Road, Shanghai 200444, China

[b]Key Laboratory of Silicate Cultural Relics Conservation (Shanghai University), Ministry of Education

AB STRACT

Small sample instance segmentation is a very challenging task, and many existing methods follow the training strategy of meta-learning which pre-train models on support set and fine-tune on query set. The pre-training phase, which is highly task related, requires a significant amount of additional training time and the selection of datasets with close proximity to ensure effectiveness. The article proposes a novel small sample instance segmentation solution from the perspective of maximizing the utilization of existing information without increasing annotation burden and training costs. The proposed method designs two modules to address the problems encountered in small sample instance segmentation. First, it helps the model fully utilize unlabeled data by learning to generate pseudo labels, increasing the number of available samples. Second, by integrating the features of text and image, more accurate classification results can be obtained. These two modules are suitable for box-free and box-dependent frameworks. In the way, the proposed method not only improves the performance of small sample instance segmentation, but also greatly reduce reliance on pre-training. We have conducted experiments in three datasets from different scenes: on land, underwater and under microscope. As evidenced by our experiments, integrated image-text corrects the confidence of classification, and pseudo labels help the model obtain preciser masks. All the results demonstrate the effectiveness and superiority of our method.

## 1. Introduction

Instance segmentation is a computer vision task that requires locating the object of interest in an image and predicting its mask and category. It is closely to semantic segmentation and object detection.

[*]Corresponding author.
han_yx@i.shu.edu.cn (Y. Han)

Compared to semantic segmentation, instance segmentation requires distinguishing different instances within the same category. Therefore, relevant methods need to capture the differences among instances of different categories and obtain the differences among instances of the same category. Compared to object detection, instance segmentation requires obtaining precise masks of objects. Thus, instance segmentation requires richer detailed information. With the development of convolutional neural networks, the task has made vigorous progress, and many methods have been widely applied in real-world scenarios, such as autonomous driving, medical image analysis, game development, and video surveillance.

Many instance segmentation methods are based on object detection by adding mask branches to object detection models. Overall, the existing methods are divided into three types. The first type is anchor-based methods, e.g. Mask R-CNN [1]. At the beginning, localization of proposals is done with RPN (Region Proposal Network). Subsequently, based on the features of proposals, classes and masks of instances can be predicted. In the type of methods, the richness of anchors and the quality of proposals greatly affect the precision of predicted masks and classes. The second type is grid-based models, representing work with SOLO [2]. SOLO is inspired by the object detection model YOLO [3] and a mask branch is added on YOLO to complete instance segmentation. Compared to anchor-based methods, the type of methods no longer slides the anchors on the image to obtain proposals, resulting in faster inference speed. While, SOLO share a common problem of missed detections with YOLO. In addition, the post-processing in the type of methods is correspondingly transformed from box-based non maximum suppression into mask-based non maximum suppression. The transformation brings a sharp increase in calculations because masks have denser representations than boxes. The third type of methods is based on queries, inspired by DETR [4], e.g. Mask2Former [5]. The type of methods benefits from the powerful ability of Transformer [6] in decoding learnable queries to obtain global relationships among instances. The type of methods is no longer limited to anchors or grids, nor does it require post-processing.

The existing models have achieved good results, but their high performance relies on large-scale data, such as the COCO dataset, which contains over 110000 training images. In many cases, only a small amount of data or limited labeled data is collected. Many recent studies have built modules based on meta-learning to solve the problem of sample scarcity, such as FGN [7] and iMTFA [8]. However, the pre-training phase of meta-learning is highly task-related and requires a significant amount of additional training time. In addition, to ensure the pre-trained weights are suitable to guide the model find the optimal solution, methods based on meta-learning also require careful selection to reduce the distance between support set and query set.

SemInst

In the paper, a novel solution for small sample instance segmentation is proposed, called SemInst (**Sem**i-supervised Instanse Segmentator with **Sem**antic Classes). The proposed method considershow to fully utilize the information of the limited labeled data, without additional datasets or annotations. The proposed method uses projective image-text to integrate visual and semantic features to enhance classification ability. And a lightweight two-stage training strategy including supervised learning and semi-supervised learning is designed to refine masks. The main contributions of the article are:

- A classification branch is added to instance segmentation model which integrates text and image information of categories to help classify objects.

- A lightweight two-stage training strategy is designed which effectively utilizes online pseudo labeling techniques to improve the utilization of unlabeled data and increase the information that the model can learn.

- The proposed method combined semantic information and two-stage training strategy outperforms many existing methods for various small sample situations.

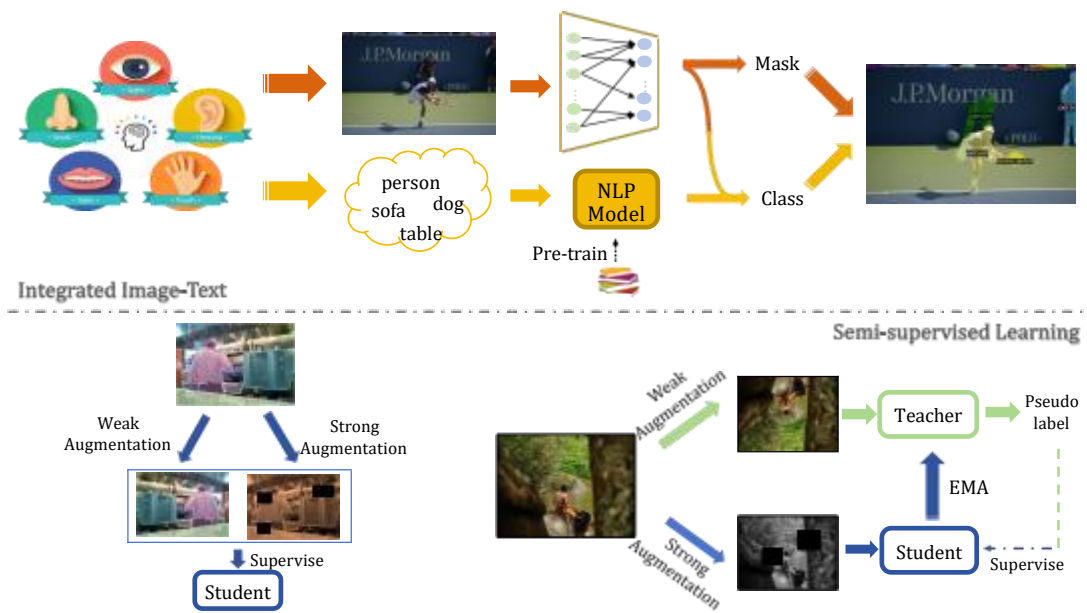

**Figure 1.** Flowchart of SemInst.

## 2. Related Works

### 2.1. Small Sample Learning

The current research on small sample learning can be divided into the following types: data augmentation,model optimization, etc. Some methods only increase data through weak transformation

Page 3 of 24

such as rotation, translation and resization. Moreover, the traditional data augmentation methods are much more complex and much more effective. For instance, Mixup [9] is a data augmentation technology originating from computer vision by randomly selecting two training samples with labels and generating a new vector as enhanced data by linear interpolation. Many augmentation methods have been developed based on Mixup [9], such as Supermix based on saliency [10] and Stylemix based on style and content separation [11]. Recently, with the development of Vision Transformer, the TokenMixup [12] has brought increasingly significant improvements in its effectiveness. But all the methods based on Mixup have an assumption that linear interpolation of features corresponds to linear interpolation of labels. Obviously, the real data distribution is not a simple linear transformation which cannot describe the real distribution well. Due to the features learned by downstream models tend to have invariance in scale, rotation, and translation, the gains brought by these augmentation methods are limited. To generate a large scale of data, the generative models can learn the data distribution with the large-scale training data, e.g., the Generative Adversarial Network (GAN) [13]. Generative Adversarial Network (GAN) requires large-scale training data makes model-based data augmentation methods not the mainstream solution for small sample learning.

The second type methods are based on model optimization. The basic idea of the type of methods is to provide pre-training weights from large scale data to easily fine-tune the target model, or to limit the solution space during training to find the optimal solution of the model. Prototypical networks [14] and matching networks [15] are based on metric learning. The features from the two networks are learned by constraining the distance between sample pairs in the model. Thus, the two networks require a lot of manual designs and experiments to determine the appropriate measurement criterion. Besides, transfer learning methods firstly train the network based on additional data and fine-tune the model on the target domain. The distance between the source domain and the target domain can affect the performance. The most representative one of transfer learning is to firstly use meta-learning to pre-train the network. The training setting of meta-learning is composed of the same training tasks on different datasets, aiming to improve the model's ability to solve specific task, such as MAML[16]. The existing meta-learning methods inject effort into the selection of base classes for pre-training and novel classes for small sample data. Moreover, meta-learning is a pre-training method with strong task correlation, which requires a significant amount of training costs and additional data. In general computer vision tasks, many prior knowledge can be shared for the transfer learning to become effective. However, in specific tasks such as material, medicine, and industrial scenarios, it is difficult to find a large amount of shared prior knowledge, so these methods are not suitable for all scenarios.

## 2.2. Semi-supervised Learning

Semi-supervised learning is a research based on the fact that although labeled data is scarce, unlabeled data is easily obtainable. Therefore, the primary goal of semi-supervised learning is how to utilize information from unlabeled data. The existing semi-supervised learning methods mainly include consistency regularization and pseudo labeling methods.

The regularization method is an idea of data augmentation, which involves adding noise to unlabeled data to minimize the difference in the model's prediction results before and after noise addition, in order to avoid overfitting and improve the model's generalization ability. For instance, Π-model [17] minimizes the difference between original result and the result of the same data after random transformations, such as dropout and random max pooling. Π-model runs each sample twice, doubling the computational cost. In order to reduce costs, Temporal Ensembling [17] maintains the exponential moving average (EMA) of model predictions overtime as the learning objective for each training sample, and only evaluate and update once in each round. Temporal Ensembling tracks the label prediction EMA of each training sample as the learning objective. However, the label prediction only changes in each round, making the method cumbersome when the training dataset is large. To overcome the problem of slow updates, Mean Teacher [18] tracks the average change of model weights instead of model outputs. Some models apply adversarial noise to the input and train the model to be robust to adversarial attacks based on adversarial training [13]. The setting is applicable to supervised learning, and Virtual Adversarial Training (VAT) [19] extends the idea to semi-supervised learning. Interpolation Consistency Training (ICT) [20] enhances the dataset by adding more data points for interpolation, and expects the prediction to be consistent with the interpolation of the corresponding labels. Following the idea of MixUp [9], ICT expects the prediction model to generate a label on the mixed sample to match the predicted interpolation of the corresponding input. Similar to VAT[19], Unsupervised Data Augmentation(UDA) [21] is particularly focused on studying how the quality of noise affects the performance of semi-supervised learning with consistency training. The consistency regularization method encourages the prediction of unlabeled data to remain consistent before and after disturbance. Trough that, the decision boundary is located in a low-density area, which effectively alleviates the phenomenon of overfitting.

The pseudo labeling method uses the teacher model to generate pseudo labels for unlabeled data, which have biases with real labels. The existing labeled data and these unlabeled data with pseudo labels are mixed together to train the student model. The teacher model and student model are trained in parallel. With meta pseudo labels [22], the teacher model is continuously adjusted based on the performance feedback of student on the labeled data. Pseudo labeling method is equivalent to entropy regularization [23], which minimizes the conditional entropy of the class probability of the unlabeled data to support low-density separation among classes. In other words, the predicted class probability is actually a measure of class overlap, and minimizing entropy is equivalent to reducing class overlap, thereby reducing density separation.

Furthermore, the combination of consistency regularization and pseudo labeling has shown good performance and become a research hotspot recently, such as MixMatch [24] and FixMatch [25].

### 2.3. Small Sample Instance Segmentation

Small sample instance segmentation is a specific task of small sample learning. FGN [7] is a small sample instance segmentation method based on meta-learning. It introduces three guidance mechanisms to Mask R-CNN, namely Attention Guided RPN, Relationship Guided Detector and Attention Guided FCN. The guidance mechanisms use the embedding of the support set as a reference for supervising the model to quickly locate and segment relevant category objects in the query set. RefT [26] uses a mask-based dynamic weighting module to enhance support features and cross attention to link target queries for better calibration. Because the support object queries encode key factors after pre-training, query features can be enhanced twice from feature level and instance level. FAPIS [27] is an anchor-free small sample instance segmentation framework to explicitly build the potential object parts shared among training object classes. FAPIS intends to score and regress the position of foreground bounding boxes, and estimate the relative importance of potential parts within each box. In addition, FAPIS specifies a new network to describe and weight the potential portion of the final instance segmentation within each detected bounding box. Overall, the development of small sample instance segmentation is closely related to meta-learning. Based on meta-learning, more effective modules are proposed for instance segmentation tasks to learn features.

### 2.4. Semantic Modules in Visual Tasks

With the development of multimodal learning, images and text, as the two most important sensory information, are combined to train the models. These methods require two encoders for text information and image information, respectively. For instance, CLIP [28] improves image recognition tasks by adding textual descriptions of images, which can be easily extended to semantic segmentation and object detection tasks. GroupViT [29] utilizes image-text pairs to supervise model training without the need for masks, allowing the model perform simple segmentation tasks. Ref.[30] uses TriNet to map the feature space to the semantic space to achieve small sample data augmentation. In many zero-shot visual tasks, semantic information is encoded and input into the network to improve the quality of image features. Many existing methods construct knowledge graphs to facilitate recognition tasks.

ADS [31] and SSR-FSD [32] are the few models for small sample object detection that use semantic embedding. However, the category information of ADS is sequentially embedded without correlation. SSR-FSD enhances the semantic space through a semantic interpretation module. OneFormer [33] adds a text encoder on top of MaskFormer[34], giving the model an input based on text conditions. In small sample instance segmentation, it is necessary to learn more accurate features due to the need for detailed masks for instance segmentation compared to small sample object detection. Although cross modal training brings more information, it also increases the complexity of training. Besides, the reasonable integratation of multimodal data still remains difficulties.

## 3. Method

The datasets for small sample learning is divided into $D_{Base}$ and $D_{Novel}$ in many meta-learning based works, where $D_{Base} \cap D_{Novel} = \emptyset$. In the setting, the model is firstly trained on $D_{Base}$ with abundant data to extract preciser features. Then, the main architecture is frozen and $D_{Novel}$ is used to fine-tune some modules in the model. Finally, the results on the classes of $D_{Novel}$ are reported. It has been proved that the training strategy can work well in many small sample tasks. But considering the efforts to collect suitable $D_{Base}$ and the expenses of training in the first stage, we don't take the training design into consideration and don't use any auxiliary annotated data to train our model. In the condition, a semi-supervised instance segmentation model called SemInst (Semi-supervised Instanse Segmentator with Semantic Classes) for small sample instance segmentation is proposed. the model improves the classification accuracy by integrating semantic information of categories, and fully utilizes unlabeled data by pseudo labeling method. By combining these two strategies, the instance segmentation performance of SemInst is improved compared to the baseline model. Figure 1 is the overall flowchart of the model. The detailed introduction are as follws.

### 3.1. Semantic Branch

Instance segmentation methods can be divided into box-free and box-dependent methods. Box-free methods are composed of two subtasks, segmenting instance masks and classifying instances. The box-dependent methods calibrate the bounding boxes of instances at the beginning and proceed with the two subtasks in each box. It can be seen that two types of methods need to classify objects. The semantic branch which fuses the text features with the image features can be used to the two types of methods for improving the precision of classification.

Semantic information is represented by natural language, so it cannot directly participate in the calculation of the model. Therefore, we use SciBert[35] which is pre-trained on large-scale texts to extract word embeddings of categories. The semantic space which is formed of categories embeddings is denoted as $W_{se} \in \mathbf{R}^{N \times d_w}$, where, $N$ is the number of categories in a dataset including background and $d_w$ is the dimension of every word embedding generated by SciBert. Considering the different dimensions of image features and word embeddings, one projection layer is used to change the dimension of $W_{se}$, which is represented as

$$W_{se-vi} = W_{se}P, \qquad (1)$$

where, $P \in \mathbf{R}^{d_w \times d}$ is the weight of projection layer. Thus, the new embeddings $W_{se-vi} \in \mathbf{R}^{N \times d}$ are obtained, where the dimension $d$ is consistent to the dimension of image features $V \in \mathbf{R}^{q \times d}$. In box-free methods, $q$ denotes the number of queries. While, $q$ denotes the number of boxes in box-dependent methods. Next, the image features $V$ and new word embeddings $W_{se-vi}$ obtained from the model are fused. Finally, the classification results are obtained after a MLP. The process mentioned above can be expressed as:

$$Pred_{cat} = Softmax(MLP(V \times W_{se-vi}^T)), \qquad (2)$$

where, $Pred_{cat} \in \mathbf{R}^{q \times N}$ is the predicted results. With the auxiliary text information of categories, the model not only learns the visual information but also learns the semantic information from scarce data.

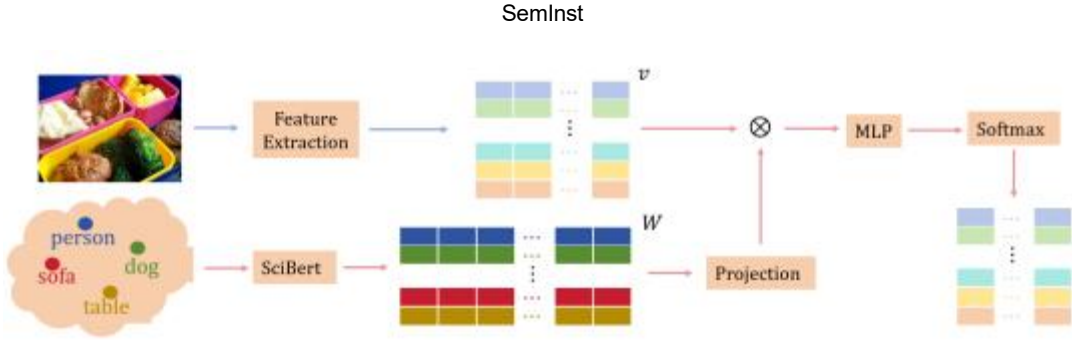

**Figure 2.** The structure of combining semantic branch and image features. The blue lines show the way image feature *v* is get, and the red lines show how the semantic branch integrate word embeddings *W* and image feature *V*. Each row in *V* represents the visual features of an instance, while each row in *W* represents the semantic representation of a category.

### 3.2. Two-stage training

To fully utilize the information of the unlabeled data, pseudo labeling technique is used as shown in Figure 1. Due to the inaccurate prediction results obtained at the beginning of training, a two-stage training strategy is designed consisted of supervised training and semi-supervised training. In the first training stage, labeled data is used to train Teacher. While, in the second training stage, Student is initialized with the model weights obtained in the supervised training stage. Teacher generates pseudo labels and Student is trained by labeled and unlabeled data with pseudo labels in semi-supervised training. In Noisy Boundary[21], the whole training is also splited into two stages. It generates pseudo labels between two stages in an offline way and mix all data to train the model from scratch. But in our method, the results of first stage are memorized and the pseudo labels are generated online.

---

**Algorithm 1:** The pseudocode of semi-supervised training for one iteration.

**Input:** Labeled data $D\_l$, unlabeled data $D\_un$, threshold $t$, keep value $r$, weight
**Output:** Teacher $M\_tea$ and Student $M\_stu$ updated after one iteration.

// Data augmentation.
1   $D\_l\_w, D\_un\_w \leftarrow WeakAug(D\_l, D\_un)$
2   $D\_l\_s, D\_un\_s \leftarrow StrongAug(D\_l, D\_un)$
   // Pseudo labels generation.
3   $pred\_D\_un \leftarrow M\_tea(D\_un\_w)$
4   $pseudo\_label \leftarrow filter(pred\_D\_un, t)$
5   $all\_D\_un \leftarrow add\_label(D\_l\_w, pseudo\_label) + add\_label(D\_l\_s, pseudo\_label)$
   // Student is trained by labeled and unlabeled data.
6   $all\_D\_l \leftarrow D\_l\_w + D\_l\_s$
7   $loss = M\_stu(all\_D\_l) + M\_stu(all\_D\_un)$
8   $loss.backward()$
   // Teacher is updated in the way of EMA.
9   $\theta_{tea} \leftarrow r\,\theta_{tea} + (1-r)\,\theta_{stu}$

---

The details are described in Algorithm 1. First, weak and strong augmentation are performed on labeled data, $D\_l$, and unlabeled data, $D\_un$. The weak augmentation contains random resization and flipping for images. The strong augmentation contains color jitter, grayscale, gaussian blur and erasing are done on the image with a certain probability. Next, only the unlabeled data after weak augmentation, $D\_un\_w$, are inputed into Teacher to generate pseudo labels. To guarantee the reliability of pseudo labels, threshold $t$ is used for filtering. Then, all labeled data, $all\_D\_l$, and unlabeled data, $all\_D\_un$, are used to supervise the training of Student. The parameters of Student, $\theta_{stu}$, are renovated by gradient descend. The parameters of Teacher, $\theta_{tea}$, are updated using EMA (Exponential Moving Average):

$$\theta_{tea} = \theta_{tea} * r + \theta_{stu} * (1-r), \tag{3}$$

where, $r$ is the value of keep rate. The update method not only retains the training parameters of supervised training, but also absorbs the influence of unlabeled data, making the parameter updates of Teacher more stable and providing more reliable pseudo labels.

## 4. Experiments

In the section, we introduce the datasets, experiment settings and implementation details. To evaluate the effectiveness of proposed methods, the results based on extensive experiments are reported, including comparison with other superior models and ablation studies.

## 4.1. Experiment setup

Our proposed small sample instance segmentation framework is applicable to not only box-dependent models but also box-free models. Thus, the box-dependent model Mask R-CNN[1] and box-free model FastInst[36] are used as the basic models, respectively. Experiments are conducted on ocean garbage dataset TrashCan[37], COCO2017[38] and 2205DSS, which is collected from duplex stainless steel with one category under microscope. The labeled samples in the three datasets are shown in Figure 3.

**Datasets.** TrashCan contains 22 categories, with 6065 images in the training set and 1147 images in the validation set. COCO2017 consists of 80 categories of images collected from natural scenes. The training set includes 118278 images, the validation set includes 5000 images, and the test set includes 40640 images. The annotations for the test set are not publicly available. 2205DSS has 42 labeled microscopic images of materials and 45 unlabeled images. In the experiments, to simulate small sample scenarios, only 0.3%, 0.5%, and 1% of the training set for COCO2017 and 1%, 2%, 3% of the training set for TrashCan are used as labeled data, respectively. Naturally, the remaining part of the training set areas unlabeled data. For the dataset of 2205DSS, we divide labeled data into training and validation sets at the ratio of 9:1. In many small sample learning works, K-shot samples for every category are extracted. It needs a lot of time to find the suitable combinations of images in which every category only has K instances. In the way, the training set after extraction lies in a limited combinations and lacks randomness in instance segmentation. In our method, since the training data are randomly selected through proportion of images, it can have good randomness and also can better maintain the distribution of the original data.

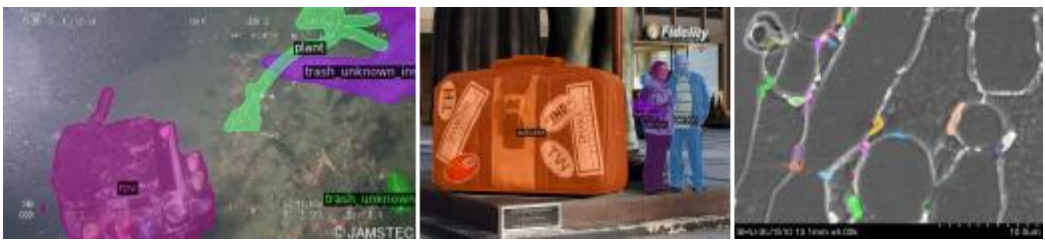

**Figure 3.** Examples. From left to right: TrashCan, COCO2017 and 2205DSS.

**Implementation Details.** The paper implements semantic branch and two-stage training based on Mask R-CNN and FastInst with the backbone of a pre-trained ResNet50. The model is trained using a optimizer of AdamW on an NVIDIA RTX3090. The initial learning rate of the model based on Mask R-CNN is 0.01, and the initial learning rate based on FastInst is 0.0001. The batch size is 4. The total training epochs of supervised training is 200, and in the semi-supervised training stage, the Student and the Teacher are trained together for 10 epochs. The keep rate of semi-supervised training *r* in Formula 3 is 0.999.

**Comparison with other models.** We provide the comparison results of SemInst based on two basic models with other known methods, including Mask R-CNN[1], FastInst[36], iFS-RCNN[39], Noisy Boundary[21] and PAIS[40]. The models can be devided into three types. Mask R-CNN and FastInst[36] represents general instance segmentation models. iFS-RCNN[39] is known as FSIS (Few Shot Instance Segmentation) method which follows the training strategy of meta-learning as mentioned before. Because we don't split datasets into $D_{Base}$ and $D_{Novel}$, iFS-RCNN is trained with all categories once for equal comparison. Noisy Boundary[21] and PAIS[40] are semi-supervised models. All the models are evaluated on validation sets of TrashCan, COCO and 2205DSS by the standard metrics of VOC challenge[41], including AP, $AP_{50}$ and $AP_{75}$.

## 4.2. Comparison Results

**Results on TrashCan.** Table 1 shows the instance segmentation performance of all the models mentioned before on the 22 classes of TrashCan at different proportions. Compared to COCO2017, TrashCan is more chanllenging for the low quality of images. From Table 1, it can be seen that our models achieve the first-best or second-best performance, except the $AP_{50}$ of 3%. Our model can attain the highest AP scores, but its competitiveness weakens from the perspective of $AP_{50}$ and $AP_{75}$. For $AP_{50}$, FastInst[36] which is the general model proposed based on large-scale data has best performance. The semi-supervised models, Noisy Boundary[21] and PAIS[40], keep a relatively stable performance. Meta-learning iFS-RCNN[39] loses its competitiveness on TrashCan without pre-training. Overall, the proposed methods can have a relatively stable high performance across all proportions and all evaluation metrics on TrashCan benchmarks. Figure 4 presents the visualization results of instance segmentation on TrashCan. iFS-RCNN, Noisy Boundary and PAIS tend to segment as many objects as they can, resulting in messy outputs. Our model based on Mask R-CNN recognize most objects than any others while our model based on FastInst gets more preciser masks.

**Results on COCO.** Table 2 shows the instance segmentation performance of all the models mentioned before on the 80 classes of COCO. Compared to TrashCan, COCO is more challenging in the aspect of having more classes. We find that iFS-RCNN[39] can't work on COCO benchmarks from its low quantitative results. While it is foreseeable because of complex classes COCO has compared to TrashCan and the $D_{Novel}$ splited from COCO. Without careful meta-learning training design, the performance of iFS-RCNN[39] is much lower than that they are reported when facing more simple training strategy and more complex dataset. Noisy Boundary[21] maintains strong competitiveness, especially in the evaluation metric of $AP_{50}$, which surpasses all comparasion models. Although both models are based on pseudo labels, PAIS[40] performs much worse than Noisy Boundary[21] on COCO. Overall, our models perform best on COCO among all comparasion models. The commonality between our model and Noisy Boundary[21] lies in that they both use labeled data to train the model first, and then generate pseudo labels to the sencond stage of training. But, Noisy Boundary[21] generate pseudo labels offline. The proposed method dynamically generate pseudo labels in the second stage of training due to the update of the Teacher. Compared to PAIS[40], the proposed method splits the training into two stages to keep the stability instead of generating pseudo labels from beginning. These efforts can help our model get rid of wrong optimization of pseudo labels and preserve the training results of labeled data. Figure 5 shows the visualization results of instance segmentation on COCO. In the first example, only our model based on Mask R-CNN segment "bottle". And in the second example, our model based on FastInst outputs the most precise masks. From the results, it can be concluded that only our model still outperform others.

**Table 1**
Instance Segmentation on TrashCan test-dev. **Bold**/underline indicate the first/second best.

| Method | 1% | | | 2% | | | 3% | | |
|---|---|---|---|---|---|---|---|---|---|
| | AP | $AP_{50}$ | $AP_{75}$ | AP | $AP_{50}$ | $AP_{75}$ | AP | $AP_{50}$ | $AP_{75}$ |
| Mask R-CNN[1] | 1.13 | 2.98 | 0.85 | 4.01 | 10.17 | 2.16 | 6.28 | 13.70 | 4.63 |
| FastInst[36] | 1.51 | 3.63 | 1.24 | 4.15 | **11.21** | 2.61 | 6.01 | **14.87** | 3.94 |
| iFS-RCNN[39] | 1.08 | 2.62 | 0.76 | 3.52 | 8.04 | 2.69 | 4.59 | 10.18 | 3.31 |
| Noisy Boundary[21] | 1.20 | 2.90 | 0.80 | 4.46 | 9.92 | **3.88** | 6.28 | 12.53 | **5.61** |
| PAIS[40] | 1.74 | **4.54** | 1.26 | 4.09 | 9.45 | 3.45 | 6.21 | 14.79 | 3.99 |
| Ours(Based on Mask R-CNN.) | 1.77 | 3.80 | 1.53 | 4.36 | 10.31 | 3.39 | **6.54** | 14.46 | 4.49 |
| Ours(Based on FastInst.) | **1.87** | 3.75 | **1.94** | 4.48 | 11.15 | 2.87 | 6.44 | 14.73 | 4.82 |

**Table 2**
Instance Segmentation on COCO test-dev. **Bold**/underline indicate the first/second best.

| Method | 1% | | | 2% | | | 3% | | |
|---|---|---|---|---|---|---|---|---|---|
| | AP | $AP_{50}$ | $AP_{75}$ | AP | $AP_{50}$ | $AP_{75}$ | AP | $AP_{50}$ | $AP_{75}$ |
| Mask R-CNN[1] | 4.21 | 10.11 | 3.03 | 6.71 | 14.79 | 5.23 | 9.53 | 19.98 | 8.17 |
| FastInst[36] | 4.41 | 8.96 | 3.96 | 6.48 | 12.65 | 6 | 9.25 | 17.85 | 8.58 |
| iFS-RCNN[39] | 0.20 | 0.51 | 0.16 | 0.31 | 0.81 | 0.22 | 0.95 | 2.23 | 0.64 |
| Noisy Boundary[21] | 4.43 | 9.51 | 3.48 | 6.68 | 13.71 | 5.83 | 9.31 | 18.29 | 8.77 |
| PAIS[40] | 2.78 | 5.94 | 2.32 | 4.33 | 9.22 | 3.53 | 6.11 | 11.83 | 5.56 |
| Ours(Based on Mask R-CNN.) | **5.03** | **11.45** | 3.75 | **7.15** | **15.68** | 5.75 | **10.26** | **20.85** | **8.86** |
| Ours(Based on FastInst.) | 4.67 | 9.13 | **4.32** | 6.76 | 12.93 | **6.37** | 9.49 | 18.17 | 8.79 |

**Results on 2205DSS.** Table 3 reports instance segmentation performance of all the models mentioned before on 2205DSS and Figure 6 shows the visualization results. It can be found that the model based on Mask R-CNN achieves the best performance, far exceeding the model based on FastInst. By summarizing the results of TrashCan and COCO, it can be concluded that the model

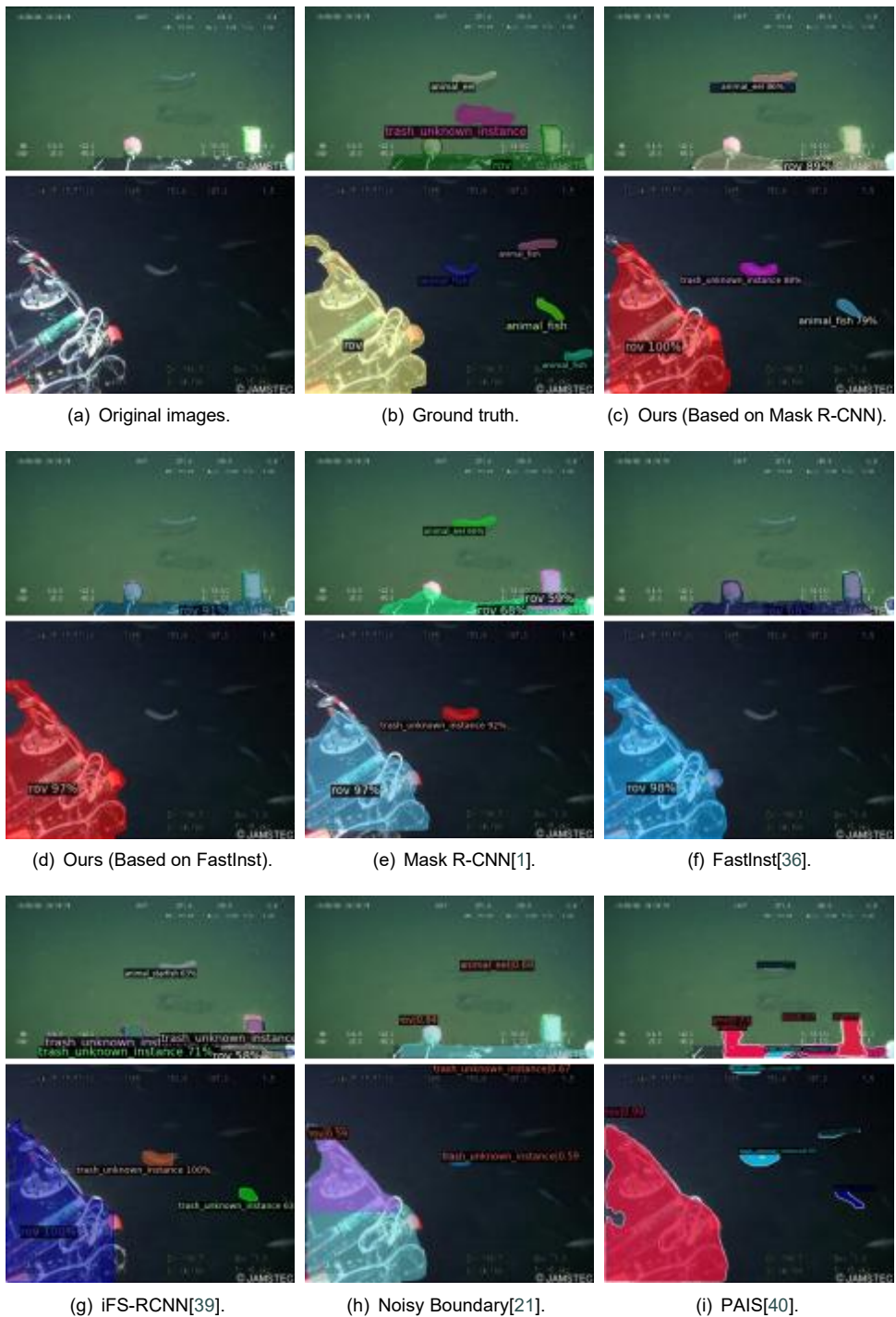

**Figure 4.** Visualized examples of results on TrashCan.

based on Mask R-CNN is more competent than the model based on FastInst for the segmentation of instances with small area. Because proposals of Mask R-CNN can effectively reduce the size of

SemInst

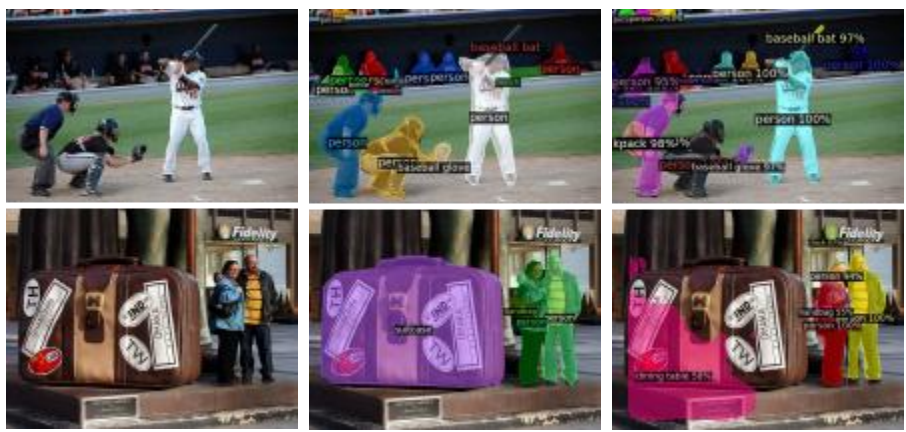

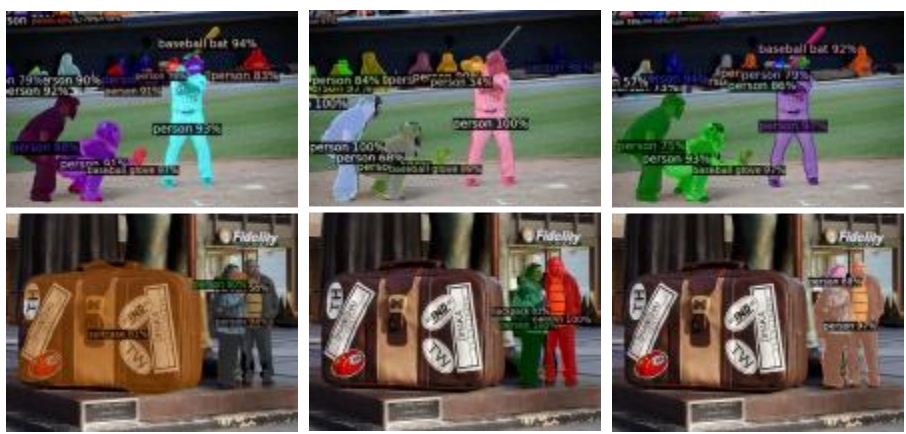

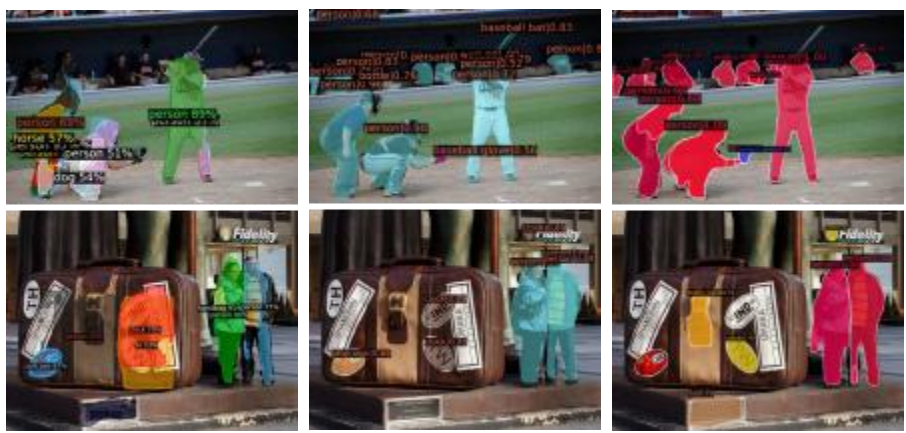

**Figure 5.** Visualized examples of results on COCO.

**Table 3**

Instance Segmentation on 2205DSS validation set. **Bold**/underline indicate the first/second best.

| Method | AP | $AP_{50}$ | $AP_{75}$ |
|---|---|---|---|
| Mask R-CNN[1] | 36.2 | 68.79 | 33.39 |
| FastInst[36] | 0.99 | 4.59 | 0.1 |
| iFS-RCNN[39] | 20.96 | 40.20 | 20.91 |
| Noisy Boundary[21] | 39.96 | 69.61 | 39.03 |
| PAIS[40] | 40.23 | 69.33 | **46.28** |
| Ours(Based on Mask R-CNN.) | **40.53** | **70.21** | <u>42.78</u> |
| Ours(Based on FastInst.) | 1.7 | 6.8 | 0.4 |

processed features, the class imbalance problem caused by the size of objects is alleviated during mask segmentation.

### 4.3. Ablation Study

In the section, we investigate the effectiveness of the proposed modules on small sample instance segmentation and the impact of hyperparameters during semi-supervised training. Specifically, based on Mask R-CNN and FastInst, we conduct ablation study of modules on the mentioned datasets, and explore the impact of the threshold *i* for pseudo labels and the ratio of labeled and unlabeled images during semi-supervised training on the 0.3% of COCO2017 training set.

**Effectiveness of Semantic Branch and Two-stage Training.** The quantitative results obtained on TrashCan, COCO2017 and 2205DSS are shown in Table 4, Table 5 and Table 6, respectively. It can be seen from the three evaluation metrics that the two additional strategies are helpful in improving the performance of the model based on box-dependent and box-free methods. Semantic branch is likely to increase $AP_{75}$ while Two-stage Training are more helpful for $AP_{50}$. It means that semantic branch are more precisely to increase the performance of predictions with higher IoU. While Two-stage Training are more evenly on all the predictions with different IoU. With the cooperation of two modules, all metrics are incresed with varying degrees compared to base model. Figure 7 and Figure 8 show the results under different combination of modules based on Mask R-CNN and FastInst to explore the effects of different modules. The experimental results of module ablation based on Mask R-CNN are shown in Figure 7. In the first examples, it can be seen that Semantic Branch helps discover unrecognized objects in the base model, while Two-stage Training helps refine the segmentation mask of category "rov". In the second examples, the Semantic Branch filters out erroneous predictions, while the Two-stage Training mined out "baseball bats" that is not detected by the base model. In Figure 8, Semantic Branch transfers the wrong classification "trash snack wrapper" into "trash unknown instance" in the first example. In the second example, the Semantic Branch mined instance "suitcase" that is not detected by the base model. After adding Two-stage Training, the model can provide a more

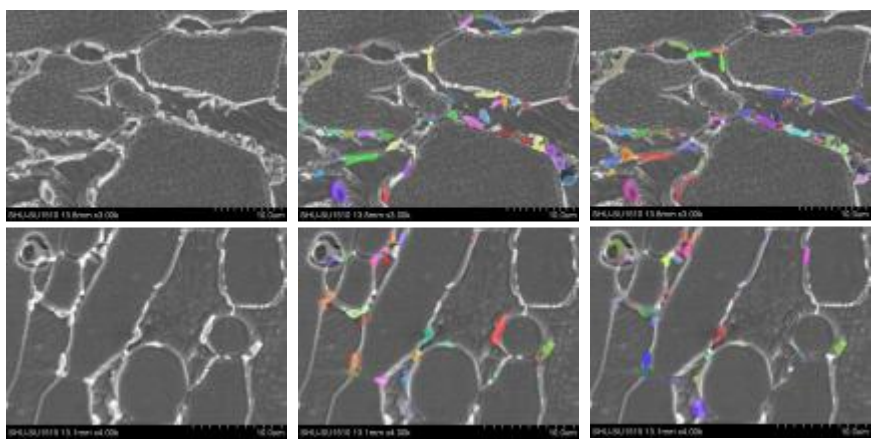

(a) Original images.  (b) Ground Truth.  (c) Ours (Based on Mask R-CNN).

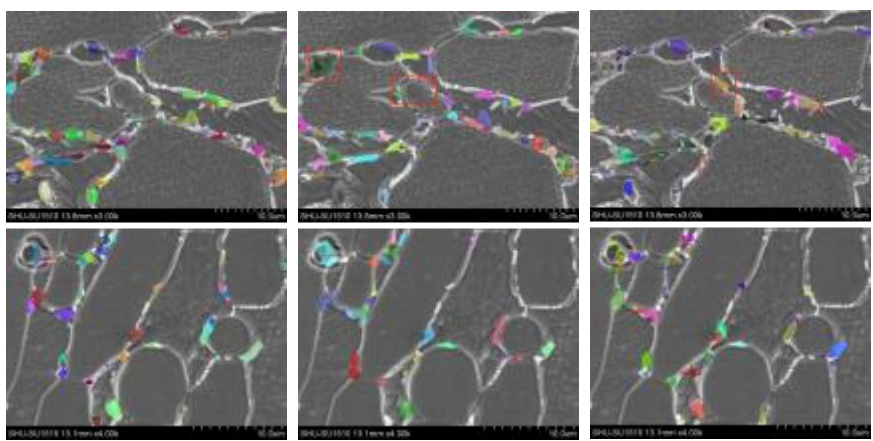

(d) Ours (Based on FastInst).  (e) Mask R-CNN[1].  (f) FastInst[36].

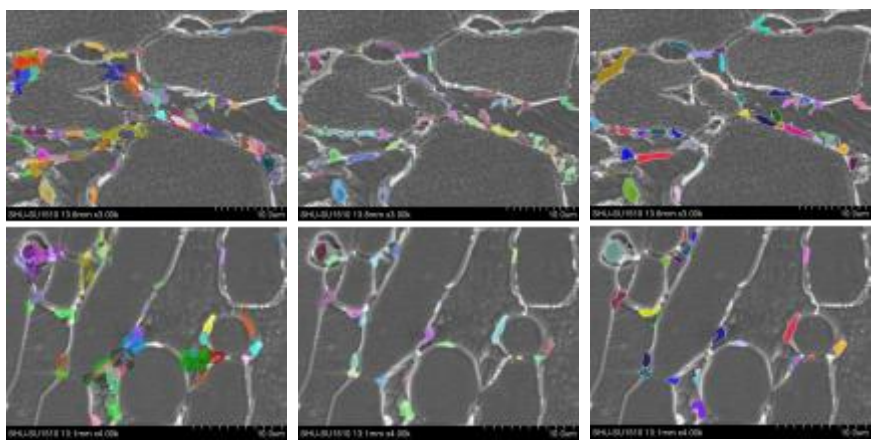

(g) iFS-RCNN[39].  (h) Noisy Boundary[21].  (i) PAIS[40].

**Figure 6.** Visualized examples of results on 2205DSS.

**Table 4**
TrashCan experimental results.

| Semantic Branch | Two-stage Training | 1% | | | 2% | | | 3% | | |
|---|---|---|---|---|---|---|---|---|---|---|
| | | AP | $AP_{50}$ | $AP_{75}$ | AP | $AP_{50}$ | $AP_{75}$ | AP | $AP_{50}$ | $AP_{75}$ |
| Based on box-dependent method, Mask R-CNN. | | | | | | | | | | |
| | | 1.13 | 2.98 | 0.85 | 4.01 | 10.17 | 2.16 | 6.28 | 13.70 | 4.63 |
| ✓ | | 1.71 | 3.68 | 1.45 | 4.32 | 10.08 | **3.45** | 6.51 | 14.46 | 4.53 |
| | ✓ | 1.21 | 3.16 | 0.92 | 4.12 | 10.36 | 2.71 | 6.34 | 13.97 | **4.73** |
| ✓ | ✓ | **1.77** | **3.80** | **1.53** | **4.36** | 10.31 | 3.39 | **6.54** | 14.46 | 4.49 |
| Based on box-free method, FastInst. | | | | | | | | | | |
| | | 1.51 | 3.63 | 1.24 | 4.15 | 11.21 | 2.61 | 6.01 | **14.87** | 3.94 |
| ✓ | | 1.83 | 3.73 | 1.75 | 4.40 | 10.95 | 2.82 | 6.29 | 14.59 | 4.56 |
| | ✓ | 1.81 | **4.42** | 1.26 | 4.32 | **11.29** | 2.80 | 6.03 | 14.82 | 4.13 |
| ✓ | ✓ | **1.87** | 3.75 | **1.94** | **4.48** | 11.15 | **2.87** | **6.44** | 14.73 | **4.82** |

**Table 5**
COCO experimental results.

| Semantic Branch | Two-stage Training | 0.3% | | | 0.5% | | | 1% | | |
|---|---|---|---|---|---|---|---|---|---|---|
| | | AP | $AP_{50}$ | $AP_{75}$ | AP | $AP_{50}$ | $AP_{75}$ | AP | $AP_{50}$ | $AP_{75}$ |
| Based on box-dependent method, Mask R-CNN. | | | | | | | | | | |
| | | 4.21 | 10.11 | 3.03 | 6.71 | 14.79 | 5.23 | 9.53 | 19.98 | 8.17 |
| ✓ | | 4.75 | 11.02 | 3.45 | 6.99 | 14.96 | **5.82** | 10.14 | 20.49 | **8.91** |
| | ✓ | 4.51 | 10.93 | 2.99 | 6.86 | 15.1 | 5.45 | 9.58 | 20.24 | 8.03 |
| ✓ | ✓ | **5.03** | **11.45** | **3.75** | **7.15** | **15.68** | 5.75 | **10.26** | **20.85** | 8.86 |
| Based on box-free method, FastInst. | | | | | | | | | | |
| | | 4.41 | 8.96 | 3.96 | 6.48 | 12.65 | 6.00 | 9.25 | 17.85 | 8.58 |
| ✓ | | 4.57 | 8.98 | 4.21 | 6.72 | 12.79 | 6.33 | 9.29 | 17.57 | 8.77 |
| | ✓ | 4.51 | 9.09 | 4.06 | 6.53 | 12.91 | 6.00 | 9.31 | 18.10 | 8.64 |
| ✓ | ✓ | **4.67** | **9.13** | **4.32** | **6.76** | **12.93** | **6.37** | **9.49** | **18.17** | **8.79** |

refined mask than only adding Semantic Branch. From the experimental results, it can be concluded that the Semantic Branch is to improve the score of correct classification and reduce the score of incorrect classification. Through the adjustment, it can help filter out incorrect segmentation and alleviate the problem of over-segmentation. The main impact of Two-stage Training is to make the masks preciser and reveal instances that are not detected in the base model. By combining the two modules, it is possible to effectively filter out erroneous predictions in the base model and generate fine masks with high confidence.

**Different thresholds for classification scores.** In semi-supervised training, pseudo labels should be filtered through threshold $i$. The paper conducts ablation experiments on the selection of thresholds as shown in Table 7. It can be seen that when the threshold is too low (0.3), incorrect pseudo labels may bring wrong guidance and the performance of the model is reduced. When the threshold is too high (0.9), few predictions participate in the training, and pseudo labels cannot be effective. From the

SemInst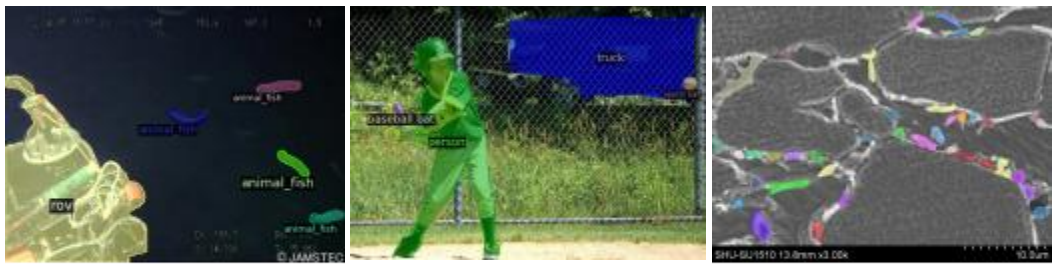
(a) Ground Truth

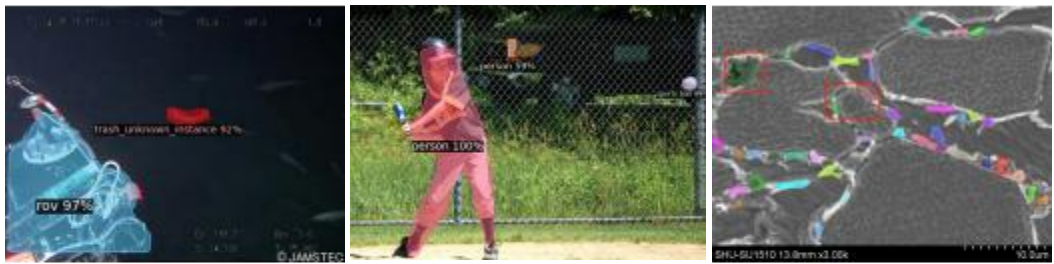
(b) Base

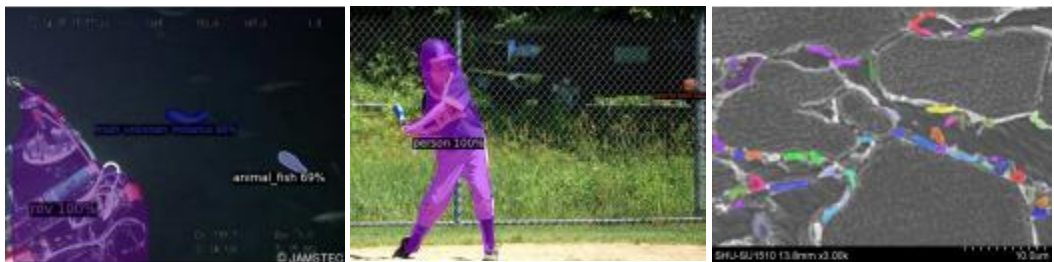
(c) +Semantic Branch

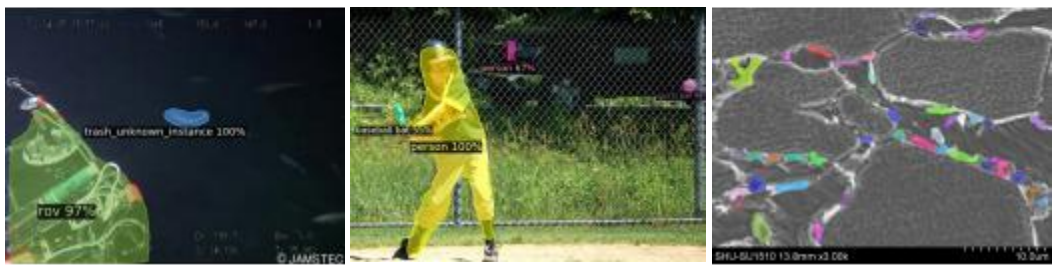
(d) + Two-stage Training

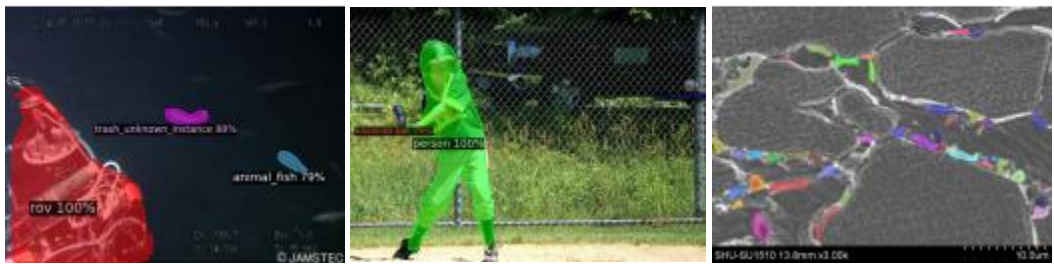
(e) + Semantic Branch and Two-stage Training

**Figure 7.** The effects of different modules based on Mask R-CNN.

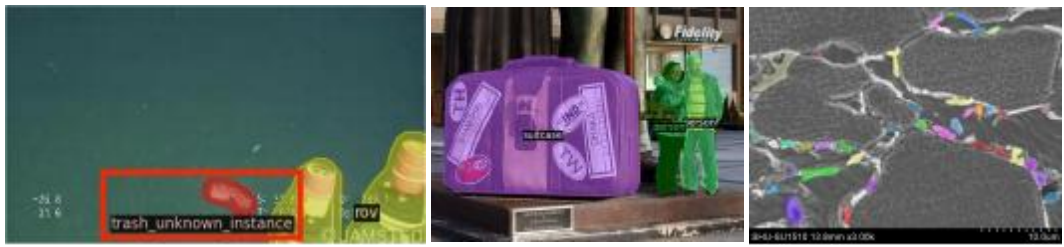

(a) Ground Truth

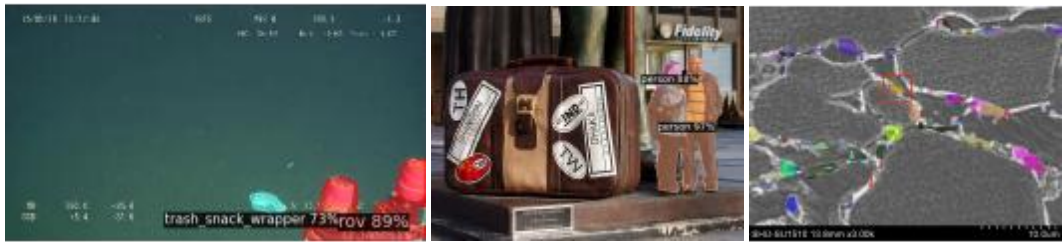

(b) Base

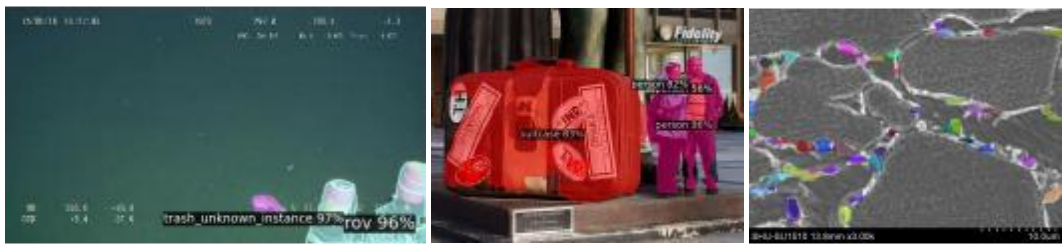

(c) +Semantic Branch

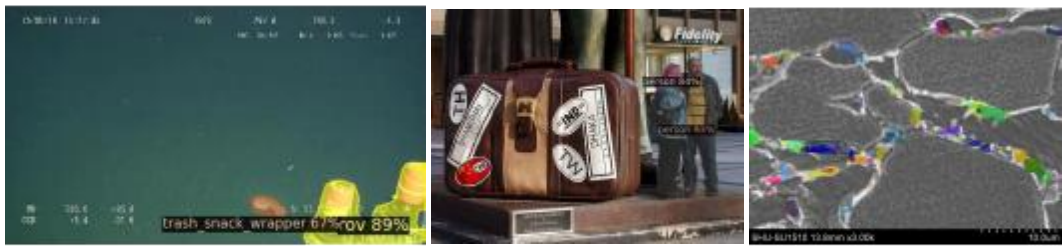

(d) + Two-stage Training

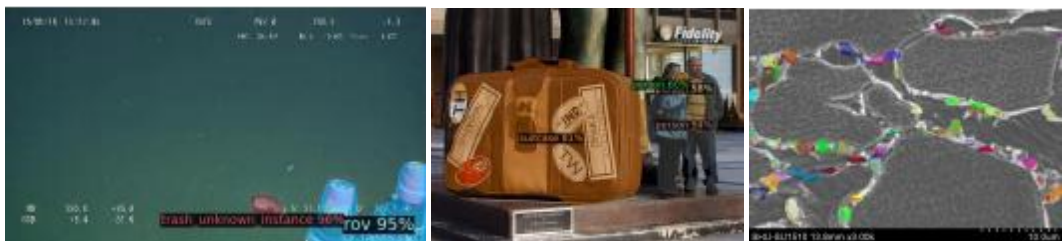

(e) + Semantic Branch and Two-stage Training

**Figure 8.** The effects of different modules based on FastInst.

**Table 6**

2205DSS experimental results.

| Semantic Branch | Two-stage Training | AP | AP$_{50}$ | AP$_{75}$ |
|---|---|---|---|---|
| Based on box-dependent method, Mask R-CNN. | | | | |
|  |  | 36.2 | 68.79 | 33.39 |
| ✓ |  | 39.84 | 69.34 | 40.91 |
|  | ✓ | 37.1 | 70 | 35.31 |
| ✓ | ✓ | **40.53** | **70.21** | **42.78** |
| Based on box-free method, FastInst. | | | | |
|  |  | 0.99 | 4.59 | 0.1 |
| ✓ |  | 1.64 | 6.38 | **0.44** |
|  | ✓ | 1.13 | 5.4 | 0.1 |
| ✓ | ✓ | **1.7** | **6.8** | 0.4 |

**Table 7**

Results of different classification score thresholds.

| ratio | Box-dependent Method | | | Box-free Method | | |
|---|---|---|---|---|---|---|
|  | AP | AP$_{50}$ | AP$_{75}$ | AP | AP$_{50}$ | AP$_{75}$ |
| 0.3 | 5.06 | 11.49 | 3.78 | 4.56 | 9 | 4.2 |
| 0.5 | 5.07 | 11.49 | 3.81 | 4.62 | 9.08 | 4.17 |
| 0.7 | 5.03 | 11.45 | 3.75 | 4.67 | 9.13 | 4.32 |
| 0.9 | 5.05 | 11.49 | 3.80 | 4.54 | 8.98 | 4.19 |

**Table 8**

Results of different ratios of labeled and unlabeled images in a batch.

| ratio | Box-dependent Method | | | Box-free Method | | |
|---|---|---|---|---|---|---|
|  | AP | AP$_{50}$ | AP$_{75}$ | AP | AP$_{50}$ | AP$_{75}$ |
| 1:1 | 5.07 | 11.49 | 3.81 | 4.67 | 9.13 | 4.32 |
| 1:2 | 5.06 | 11.48 | 3.76 | 4.57 | 9 | 4.2 |
| 1:3 | 4.81 | 11.26 | 3.39 | 4.56 | 9.02 | 4.11 |
| 1:4 | 4.81 | 11.25 | 3.4 | 4.62 | 9.08 | 4.17 |

quantitative results, it can be seen that threshold of 0.7 has better effect for box-free method, while threshold of 0.5 is more suitable for box-dependent method.

**Different ratios of labeled and unlabeled images in a batch.** In the semi-supervised training stage, the ratio of labeled and unlabeled data can also affect the performance of the model, as shown in Table 8. It can be seen that regardless of whether it is based on boxes, setting the same number of labeled and unlabeled data in each batch can achieve better performance.

## 5. Conclusion

In the paper, we delve into the problem of visual task instance segmentation when faced with sample scarcity. Semantic Branch and Two-stage Training are proposed to improve the performance of general instance segmentation methods in data-scarce scenario. The proposed method, named SemInst, achieves comparable performance with integrated image-text and semi-supervised training. As evidenced by the experimental results, Semantic Branch and Two-stage training play different role in improving the ability of instance segmentation. Specifically, integrated image-text makes the model obtain higher confidence for correct classification and reduce the confidence for wrong classification. Besides, Two-stage Training which employs consistency regularization and pseudo labeling helps the model obtain preciser masks as shown in the visualization of results. Without carefully designed training processes, we hope our work can offer some references and inspiration to foucus on how to fully extract information from scarce data. Despite its simplicity, SemInst still achieves the first-best or sencond-best performance on Trashcan, COCO2017 and 2205DSS, which indicates its effectiveness.

## 6. Acknowledgments